\documentclass[11pt]{article}
\usepackage{colacl}

\def\Nat{{\Bbb N}}

\makeatletter

\font\tenmsa=msam10
\font\sevenmsa=msam7
\font\fivemsa=msam5
\font\tenmsb=msbm10
\font\sevenmsb=msbm7
\font\fivemsb=msbm5
\newfam\msafam
\newfam\msbfam
 \textfont\msafam=\tenmsa  \scriptfont\msafam=\sevenmsa
 \scriptscriptfont\msafam=\fivemsa
 \textfont\msbfam=\tenmsb  \scriptfont\msbfam=\sevenmsb
  \scriptscriptfont\msbfam=\fivemsb

\def\hexnumber@#1{\ifcase#1 0\or1\or2\or3\or4\or5\or6\or7\or8\or9\or
	A\or B\or C\or D\or E\or F\fi }

\font\teneuf=eufm10
\font\seveneuf=eufm7
\font\fiveeuf=eufm5
\newfam\euffam
\textfont\euffam=\teneuf
\scriptfont\euffam=\seveneuf
\scriptscriptfont\euffam=\fiveeuf
\def\frak{\ifmmode\let\next\frak@\else
\def\next{\errmessage{Use \string\frak\space only in math mode}}\fi\next}
\def\goth{\ifmmode\let\next\frak@\else
\def\next{\errmessage{Use \string\goth\space only in math mode}}\fi\next}
\def\frak@#1{{\frak@@{#1}}}
\def\frak@@#1{\fam\euffam#1}
\edef\msa@{\hexnumber@\msafam}
\edef\msb@{\hexnumber@\msbfam}

\mathchardef\diagup="3\msb@1E
\mathchardef\diagdown="3\msb@1F

\def\Bbb{\ifmmode\let\next\Bbb@\else
 \def\next{\errmessage{Use \string\Bbb\space only in math mode}}\fi\next}
\def\Bbb@#1{{\Bbb@@{#1}}}
\def\Bbb@@#1{\fam\msbfam#1}

\makeatother

\def\eq{\mathrel{\approx}}

\newcommand{\tmpred}[1]{\mathrel{\triangleleft_{#1}}}
\newcommand{\tmbp}[1]{\mathrel{\bar{\triangleleft}_{#1}}}
\newcommand{\tmrtp}[1]{\mathrel{\triangleleft^*_{#1}}}
\newcommand{\tmptp}[1]{\mathrel{\triangleleft^+_{#1}}}

\newcommand{\tmileft}{\tmpred{1}}
\newcommand{\tmbleft}{\tmbp{1}}

\newcommand{\tmleft}{\tmptp{1}}

\newcommand{\tmparent}{\tmpred{2}}
\newcommand{\tmbdom}{\tmbp{2}}

\newcommand{\tmdom}{\tmrtp{2}}
\newcommand{\tmpdom}{\tmptp{2}}

\newcommand{\tmiabove}{\tmpred{3}}
\newcommand{\tmbabove}{\tmbp{3}}

\newcommand{\tmabove}{\tmrtp{3}}
\newcommand{\tmpabove}{\tmptp{3}}

\newcommand{\sym}[1]{\mbox{`$#1$'}}
\newcommand{\power}{\mbox{\boldmath $\sP$}}
\newcommand{\pair}[2]%
  {\left\langle%\vphantom{\raise1pt\hbox{$#1#2$}}
#1,#2\right\rangle}
\newcommand{\Tup}[1]%
  {\left\langle%\vphantom{\raise1pt\hbox{$#1$}}
#1\right\rangle}
\newcommand{\tup}[1]%
  {\langle{#1}\rangle}
\newcommand{\sA}{{\cal A}}
\newcommand{\sT}{{\cal T}}
\newcommand{\eps}{{\ifmmode{\mathord{\varepsilon}}%
\else{$\mathord{\varepsilon}$}\fi}}
\def\defiff{\mathrel{\buildrel \rm def \over \Longleftrightarrow}}
\def\defeq{\mathrel{\buildrel \rm def \over =}}
\newcommand{\tfin}{\mbox{ finite }}

\newcommand{\tand}{\mbox{ and }}

\newcommand{\tow}{\mbox{ otherwise }}
\newcommand{\f}[1]{{\mathchoice{\mathop{\mbox{\rm #1}}}%
{\mathop{\mbox{\rm #1}}}{\mathop{\mbox{\scriptsize\rm #1}}}%
{\mathop{\mbox{\scriptsize\rm #1}}}}}
\def\restrchr{\mathord{|}}
\newcommand{\restr}[1]{\restrchr\lower.75ex\hbox{$#1$}}
\newcommand{\set}[1]{\{#1\}}
\newcommand{\sP}{{\cal P}}
\def\sfT{{\sf T}}                       
\newcommand{\xth}[1]{\mbox{$#1^{\rm th}\;$}}
\newcommand{\bm}[1]{{\ifmmode{\mathchoice{\mathord{\mbox{\boldmath $#1$}}}%
{\mathord{\mbox{\boldmath $#1$}}}%
{\mathord{\mbox{\scriptsize\boldmath $#1$}}}%
{\mathord{\mbox{\scriptsize\boldmath $#1$}}}}%
\else{$\mathchoice{\mathord{\mbox{\boldmath $#1$}}}%
{\mathord{\mbox{\boldmath $#1$}}}%
{\mathord{\mbox{\scriptsize\boldmath $#1$}}}%
{\mathord{\mbox{\scriptsize\boldmath $#1$}}}$}%
\fi}}
\let\limp=\rightarrow
\let\liff=\leftrightarrow

\newcommand{\littletree}[3]{{\scriptstyle%
\phantom{\diagup}_{\raisebox{-1ex}{$\scriptstyle{#2}$}}%
\diagup^{\raisebox{0.5ex}{$\scriptstyle{#1}$}}%
\diagdown_{\raisebox{-1ex}{$\scriptstyle{#3}$}}}}

\title{A Descriptive Characterization of Tree-Adjoining Languages\\
(Full Version)
\thanks{This is the full version of a paper to appear in the proceedings of
COLING-ACL'98 as a project note~(Rogers, 1998).}
}

\author{James Rogers\\Dept. of Computer Science\\Univ. of Central Florida\\
        Orlando, FL 32816-2362, USA\\
        \texttt{jrogers@cs.ucf.edu}}

\begin{document}
\maketitle
\bibliographystyle{acl}
\nocite{rogers98}

\begin{abstract}
  Since the early Sixties and Seventies it has been known that the regular and
context-free languages are characterized by definability in the monadic
second-order theory of certain structures.  More recently, these descriptive
characterizations have been used to obtain complexity results for constraint-
and principle-based theories of syntax and to provide a uniform model-theoretic
framework for exploring the relationship between theories expressed in
disparate formal terms.  These results have been limited, to an extent, by the
lack of descriptive characterizations of language classes beyond the
context-free.  Recently, we have shown that tree-adjoining languages (in a
mildly generalized form) can be characterized by recognition by automata
operating on three-dimensional tree manifolds, a three-dimensional analog of
trees.  In this paper, we exploit these automata-theoretic results to obtain a
characterization of the tree-adjoining languages by definability in the monadic
second-order theory of these three-dimensional tree manifolds.  This not only
opens the way to extending the tools of model-theoretic syntax to the level of
TALs, but provides a highly flexible mechanism for defining TAGs in terms of
logical constraints.
\end{abstract}
\section{Introduction}
In the early Sixties B\"uchi~\shortcite{buchi60} and Elgot~\shortcite{elgot61}
established that a set of strings was regular iff it was definable in the weak
monadic second-order theory of the natural numbers with successor (wS1S).  In
the early Seventies an extension to the context-free languages was obtained by
Thatcher and Wright~\shortcite{ThaWri68} and Doner~\shortcite{doner70} who
established that the CFLs were all and only the sets of strings forming the
yield of sets of finite trees definable in the weak monadic second-order theory
of multiple successors (wSnS).  These descriptive characterizations have
natural application to constraint- and principle-based theories of syntax.
We have employed them in exploring the language-theoretic complexity of
theories in GB~\cite{rogers94,rogers95} and GPSG~\cite{rogers97} and have
used these model-theoretic interpretations as a uniform framework in which to
compare these formalisms~\cite{rogers96a}.  They have also  provided a
foundation for an 
approach to principle-based parsing via compilation into tree-automata
\cite{MorCor97}.  Outside the realm of Computational Linguistics, these results
have been employed in theorem proving with applications to program and
hardware verification~\cite{HeJeJo96,BiKlRa96,KeMaMeGs97}.
The scope of each of these applications is limited, to some extent, by the fact
that there are no such descriptive characterizations of classes of languages
beyond the context-free.  As a result, there has been considerable interest in
extending the basic results~\cite{moenn97,volger97} but, prior to the work
reported 
here, the proposed extensions have not preserved the simplicity
of the original results.  

Recently, in \cite{rogers97b}, we introduced a class of
labeled three-dimensional tree-like structures (three-dimensional tree
manifolds---$3$-TM) which serve simultaneously as the derived and
derivation 
structures of Tree Adjoining-Grammars (TAGs) in exactly the same way that
labeled 
trees can serve as both derived and derivation structures for CFGs.  We
defined a class of automata over these structures that are a natural
generalization of tree-automata (which are, in turn, an analogous
generalization of ordinary finite-state automata over strings) and showed that
the class of tree manifolds recognized by these automata are exactly the class
of tree manifolds generated by TAGs if one relaxes the usual requirement that
the labels of the root and foot of an auxiliary tree and the label of the node
at which it adjoins all be identical.  

Thus there are analogous classes of
automata at the level of labeled three-dimensional tree manifolds, the level of
labeled trees and at the level of strings (which can be understood as two- and
one-dimensional tree manifolds) which recognize sets 
of structures that yield, respectively, the TALs, the CFLs, and the regular
languages.  Furthermore, the nature of the generalization between each level
and the next is simple enough that many results lift directly from one level to
the next.  In particular, we get that the recognizable sets at each level are
closed under union, intersection, relative complement, projection,
cylindrification, and determinization and that emptiness of the recognizable
sets is decidable.    
These are exactly the properties one
needs to establish that recognizability by the automata over a class of
structures characterizes satisfiability of monadic second-order formulae in the
language appropriate for that class.  Thus, just as the proofs of closure
properties lift directly from one level to the next, Doner's and Thatcher and
Wright's proofs that the recognizable sets of trees are characterized by
definability in wSnS lift directly to a proof that the recognizable sets of
three-dimensional tree manifolds are characterized by definability in their
weak monadic second-order theory (which we will refer to as wSnT3).

In this paper we carry out this program.  In the next three sections we
introduce $3$-TMs and our uniform notion of automaton over tree manifolds
of arbitrary (finite) dimension and sketch, as an example, proofs of
closure under determinization, projection and cylindrification that are
independent of the dimensionality.  In Sections~\ref{sec.lang}
and~\ref{sec.def} we introduce wSnT3, the weak monadic second-order theory of
$n$-branching $3$-TM, and sketch the proof that the sets recognized by $3$-TM
automata are exactly the sets definable in wSnT3.  This, when coupled with the
characterization of TALs in Rogers~\shortcite{rogers97b}, gives us our
descriptive characterization of TALs: a set of strings is generated by a TAG
(modulo the generalization of Rogers~\shortcite{rogers97b}) iff it is the
(string) yield of a set of $3$-TM definable in wSnT3. Finally, in
Section~\ref{sec.apps} we look at how working in wSnT3 allows a potentially
more natural means of defining TALs and, in particular, a simplified treatment
of constraints on modifiers in TAGs.

\section{Tree Manifolds}
Tree manifolds are a generalization to arbitrary dimensions of Gorn's
\emph{tree domains}~\cite{gorn65}.  A tree domain is a set of node address
drawn 
from $\Nat^*$ (that is, a set of strings of natural numbers) in which
$\varepsilon$ is the address of the root and the children of a node at address
$w$ occur at addresses $w0, w1, \ldots$, in left-to-right order.  To be well
formed, a tree domain must 
be downward closed wrt to domination, which corresponds to being prefix closed,
and left sibling closed in the sense that if $wi$ occurs then so does $wj$ for
all $j<i$.  In generalizing these, we can define a one-dimensional analog as
\emph{string domains}: downward closed sets of natural numbers interpreted as
string addresses.  From this point of view, the address of a node in a tree
domain can be understood as the sequence of string addresses one follows in
tracing the path from the root to that node.  If we represent $\Nat$ in unary
(with $n$ represented as $1^n$) then the downward closure property of string
domains becomes a form of prefix closure analogous to downward closure wrt
domination in tree domains, tree domains become sequences of sequences of
\sym{1}s, and the left-closure property of tree domains becomes a prefix
closure property for the embedded sequences.

Raising this to higher dimensions, we obtain, next, a class of structures in
which each node expands into a (possibly empty) tree.  A,
\emph{three-dimensional tree manifold} ($3$-TM), then, is set of sequences of
tree addresses (that is, addresses of nodes in tree domains) tracing the paths
from the root of one of these structures to each of the nodes in it.  Again
this must be 
downward closed wrt domination in the third dimension, equivalently wrt prefix,
the sets of tree addresses labeling the children of any node must be downward
closed wrt domination in the second dimension (again wrt to prefix), and the
sets of string addresses labeling the children of any node in any of these
trees must be downward closed wrt domination in the first dimension (left-of,
and, yet again, prefix).\footnote{While this clearly iterates to obtain tree
manifolds of any finite dimension, we are concerned only with the first three
dimensions (four, counting points---zero-dimensional tree manifolds).}  Thus
$3$-TM, tree domains ($2$-TM), and string domains ($1$-TM) can be defined
uniformly as 
$d^{th}$-order sequences of \sym{1}s which are hereditarily prefix closed.  We
will denote the set of all $3$-TM as $\Bbb{T}^d$, so
$\Bbb{T}^1\subseteq\power(1^*)$ is the set of all string domains,
$\Bbb{T}^2\subseteq\power((1^*)^*)$ the set of all tree domains, and
$\Bbb{T}^3\subseteq\power(((1^*)^*)^*)$ the set of all $3$-TM, where
$\power(S)$ is 
the power set of $S$.\footnote{In~\cite{rogers97b} we constructed
tree-manifolds 
from headed strings in order to obtain a unique tree as the two-dimensional
yield of a $3$-TM.  Here we treat this as a matter of interpretation, keeping
the simple notion of tree-manifold and moving the issue of headedness into the
relational structures we build on them.}

For any alphabet $\Sigma$, a \emph{$\Sigma$-labeled $d$-dimensional tree
manifold} is a pair 
$\pair{T}{\tau}$ where $T$ is a $d$-TM and $\tau:T\to\Sigma$ is an assignment
of labels in $\Sigma$ to the nodes in $T$.  We will denote the set of all
$\Sigma$-labeled $d$-TM as $\Bbb{T}^d_\Sigma$.

\section{Tree Manifold Automata}
Mimicking the development of tree manifolds, we can define automata over
labeled $3$-TM as a generalization of automata over labeled tree domains which,
in turn, can be understood as an analogous generalization of ordinary 
finite-state automata over strings (labeled string domains).  A $d$-TM
automaton with state set $Q$ and alphabet $\Sigma$ is a finite set:
\[
\sA^d\subseteq \Sigma\times Q\times \Bbb{T}_Q^{d-1}.
\]
The interpretation of a tuple $\tup{\sigma,q,\sT}\in\sA^d$ is that if a node
of a $d$-TM is labeled $\sigma$ and $\sT$ encodes the assignment of states to
its children, then that node may be assigned state $q$.\footnote{This is a
``bottom-up'' interpretation.  There is an analogous ``top-down''
interpretation, but for all $d\geq2$ automata that are deterministic under the
top-down interpretation are strictly weaker than those that are
non-deterministic, while those that are deterministic under the bottom-up
interpretation are equivalent to the non-deterministic variety.  It should be
emphasized that the only place the distinction between top-down and bottom-up
arises is in the definition of determinism.  These automata are interpreted
purely declaratively, as licensing assignments of states to nodes.}  A
\emph{run} of an $d$-TM automaton $\sA$ on a $\Sigma$-labeled $d$-TM
$\sT=\pair{T}{\tau}$ is an assignment $r:T\to Q$ of states in $Q$ to nodes in
$T$ in which each assignment is licensed by $\sA$.  Note
that this implies that a maximal node (wrt to the major dimension) labeled
$\sigma$ may be assigned state $q$ only if there is a tuple
$\tup{\sigma,q,\eps}\in \sA^d$ where $\eps$ is the empty $(d-1)$-TM.
If we let $Q_0\subseteq Q$ be any set of
\emph{accepting states}, then the set of (finite) $\Sigma$-labeled $d$-TM
recognized by $\sA$, relative to $Q_0$, is that set for which there is a run of
$\sA$ that assigns the root a state in $Q_0$:
\[
\begin{array}{ll}
\multicolumn{2}{l}{
\sA(Q_0)\defeq}\\
\multicolumn{2}{l}{\quad\{\sT=\pair{T}{\tau}\mid T \tfin \tand}\\
\quad&
\exists r:T\to Q \mbox{ such that }\\
&\quad r(\eps)\in Q_0 \tand\mbox{ for all } s\in T\\
&\quad \left\langle
\tau(s),r(s),\pair{T}{r}\restr{\f{Ch}(T,s)}\right\rangle\in\sA \,\}
\end{array}
\]
where 
$\f{Ch}(T,s)=\set{w\in\Bbb{T}^{(d-1)}\mid s\cdot \tup{w}\in T}$
 and
\[
\begin{array}{ll}
\multicolumn{2}{l}{\pair{T}{r}\restr{\f{Ch}(T,s)}=}\\
\quad&\pair{\f{Ch}(T,s)}{\set{w\mapsto r(s\cdot\tup{w})\mid w\in\f{Ch}(T,s)}}.
\end{array}%
\footnote{In general, we will employ $w$ and $s$ in this manner where $w$
denotes a sequence of some order and $s$ denotes a sequence of sequences of
the order of $w$ (i.e., a sequence of the next higher order). Concatenation
will always be interpreted as an operation on sequences of the same order.
Thus, $s\cdot\tup{w}$ is a sequence of sequences in which the last sequence is
$w$.  We will also use $t$ and $v$ as we use $s$ and $w$, and will employ $p$
for sequences of the next higher order than $s$ and $t$ when needed.}
\]
A set of $d$-TM is \emph{recognizable} iff it is $\sA(Q_0)$ for some $d$-TM
automaton $\sA$ and set of accepting states $Q_0$.

\section{Uniform Properties of Recognizable Sets}
The strength of the uniform definition of $d$-TM automata is that many, even
most, properties of the sets they recognize can be proved
uniformly---independently of their dimension. 
For instance, let us say that the \emph{depth} of a TM is the length of the
longest sequence it includes (just the length of the top level sequence,
independent of the length of the sequences it may contain).  The
\emph{branching factor} of a TM \emph{at a given dimension} is the maximum
depth of the structures it contains in that dimension.  The (overall)
\emph{branching factor} of a $d$-TM is the maximum of its branching factors at
all dimensions strictly less than 
$d$.  For a $3$-TM, then, the branching factor
is the larger of the maximum depth of the trees it contains and the maximum
length of the strings it contains.  A TM is \emph{$n$-branching} iff its
branching factor is no greater than $n$.
We will
denote the set of all $\Sigma$-labeled, $n$-branching, $d$-TM as
$\Bbb{T}_\Sigma^{n,d}$.  A $d$-TM automaton is \emph{deterministic} with
respect to a branching factor $n$ (in the bottom-up sense) iff
\[
(\forall \sigma\in\Sigma, \sT\in\Bbb{T}_Q^{n,d-1})
(\exists! q\in Q)[\tup{\sigma,q,\sT}\in\sA].%
\footnote{The quantifier $\exists!$ should be read ``exists exactly one''.}
\]

It is easy to show, using a standard subset-construction, that (bottom-up)
determinism does not effect the recognizing 
power of $d$-TM automata of any dimension.
Given $\sA\subseteq \Sigma\times Q\times\Bbb{T}_Q^{n,d-1}$, let
\begin{eqnarray*}
\lefteqn{\hat{\sA}\subseteq
\Sigma\times\power(Q)\times\Bbb{T}_{\power(Q)}^{n,d-1}}\\
&\defeq&
\{ \tup{\sigma,Q_1,\pair{T}{\tau'}} \mid\\
&&\quad  Q_1\subseteq Q,\; \tau':T\to\power(Q),\\
&&\quad q\in Q_1\iff\\
&&\qquad (\exists \tau:T\to Q)[\\
&&\qquad\quad \tup{\sigma,q,\pair{T}{\tau}}\in\sA\land\\
&&\qquad\quad (\forall x\in T)[\tau(x)\in\tau'(x)]\;],\\
\hat{Q}_0&\defeq&\set{Q_i\subseteq Q\mid Q_i\cap Q_0\neq \emptyset}.
\end{eqnarray*}
It is easy to verify that $\hat{\sA}$ is deterministic and that 
$\hat{\sA}(\hat{Q}_0)=\sA(Q_0)$.  More importantly, while the dimension of the
TM automaton parameterizes the type of the objects manipulated by the proof,
it has no effect on the way in which they are manipulated---the proof itself is
essentially independent of the dimension.

Proof of closure of recognizable sets under projection and cylindrification is
even easier.  A \emph{projection} is any (usually many-to-one) surjective map
from one alphabet onto another.  A \emph{cylindrification} is an ``inverse''
projection.  Let $\pi:\Sigma\to\Sigma'$ be any projection, 
$\sT=\pair{T}{\tau}$ a $\Sigma$-labeled $d$-TM and $\sA$ an automaton over
$\Sigma$-labeled $d$-TM.  Then 
$\pi(\sT)\defeq\pair{T}{\pi\circ\tau}$ and
\[
\pi(\sA)\defeq\set{\tup{\pi(\sigma),q,\sT}\mid\tup{\sigma,q,\sT}\in\sA}.
\]
It is easy to see that 
\[
\sT\in\sA(Q_0)\iff\pi(\sT)\in\pi(\sA)(Q_0).
\]
Similarly, if $\sA\subseteq \Sigma'\times Q\times \Bbb{T}_Q^{(d-1)}$
let 
\[
\pi^{-1}(\sA)\defeq\set{\tup{\sigma,q,\sT}\mid
\tup{\pi(\sigma),q,\sT}\in\sA}.
\]
Then $\pi(\sT)\in\sA(Q_0)\iff\sT\in\pi^{-1}(\sA)(Q_0)$.

Similar uniform proofs can be obtained for closure of recognizable sets under
Boolean operations and for decidability of emptiness.

\section{wSnT3}\label{sec.lang}
We are now in a position to build relational structures on $d$-dimensional tree
manifolds.  Let $T^d_n$ be the \emph{complete $n$-branching $d$-TM}---that
in which every point has a child structure that has depth $n$ in all its
($d-1$) dimensions.  Let
\[
\sfT^3_n\defeq\tup{T^3_n,\tmileft,\tmparent,\tmiabove}
\]
where, for all $x,y\in T^3_n$:
\begin{eqnarray*}
x\tmiabove y &\defiff& y=x\cdot\tup{s}\\
x\tmparent y &\defiff& x=p\cdot\tup{s}\tand y=p\cdot\tup{s\cdot\tup{w}}\\
x\tmileft y &\defiff& x=p\cdot\tup{s\cdot\tup{w}} \tand\\
&&\quad                        y=p\cdot\tup{s\cdot\tup{w\cdot v}}
\end{eqnarray*}
where $p\in((1^*)^*)^*,\; s\in(1^*)^*,\; w\in 1^*,\; v\in 1^+$
(which is to say that 
$x\tmpred{i} y$ iff $x$ is the immediate predecessor of
$y$ in the \xth{i}-dimension).

The \emph{weak monadic second-order language} of $\sfT^3_n$ includes constants
for each of the relations (we let them stand for themselves), the usual logical
connectives, quantifiers and grouping symbols, and two countably infinite sets
of variables, one ranging over individuals (for which we employ
lowercase) and one ranging over \emph{finite} subsets (for which
we employ uppercase).  If $\varphi(x_1,\ldots,x_n,X_1,\ldots,X_m)$ is a formula
of this language with free variables among the $x_i$ and $X_j$, then we will
assert that it is satisfied in $\sfT^3_n$ by an assignment $\bm{s}$ (mapping
the \sym{x_i}s to individuals and \sym{X_j}s to finite subsets) with the
notation
\[
\sfT^3_n\models\varphi\;[\bm{s}].
\]
A \emph{sentence} is a formula with no free variables---formulae for which
truth in $\sfT^3_n$ is not contingent on an assignment.  The set of all
sentences of this language that are satisfied by $\sfT^3_n$ is the
\emph{weak monadic second-order theory of $\sfT^3_n$}, denoted
wSnT3.\footnote{wS1T1 is equivalent to wS1S in the sense of
interinterpretability, as is wS1Td for all $d$.  wSnT2 is interinterpretable
with wSnS for all $n\geq 2$.}

\section{Definability in wSnT3}\label{sec.def}
A set $\Bbb{T}$ of $\Sigma$-labeled $3$-TM is definable in wSnT3 iff there
is a formula 
$\varphi_{\raisebox{-0.5ex}{$\scriptstyle\Bbb{T}$}}%
(X_T,X_\sigma)_{\sigma\in\Sigma}$,
with free 
variables among $X_T$ (interpreted as the domain of a tree) and $X_\sigma$ for
each $\sigma\in\Sigma$ (interpreted as the set of $\sigma$-labeled points in
$T$), such that
\[
\begin{array}{lr}
\multicolumn{2}{l}{\pair{T}{\tau}\in\Bbb{T}\iff }\\
&\sfT^3_n\models\varphi_{\raisebox{-0.5ex}{$\scriptstyle\Bbb{T}$}}\;
[X_t\mapsto T, X_\sigma\mapsto\set{p\mid\tau(p)=\sigma}].
\end{array}
\]

It should be reasonably easy to see how any recognizable set can be defined in
this way.  Suppose the \xth{i} tuple of $3$-TM automaton $\sA$ is
$\tup{a,0,\littletree{1}{0}{1}}$.  A local (depth one in its major dimension)
$3$-TM (labeled with both $\Sigma$ and $Q$) is compatible with this iff its
root satisfies 
\[
\begin{array}{ll}
\multicolumn{2}{l}{\varphi_{i}(x)\equiv}\\
\multicolumn{2}{l}{\;(\exists x_1,x_2,x_3)[
X_T(x_1)\land X_T(x_2)\land X_T(x_3)\land}\\
& X_a(x)\land X_0(x)\land X_1(x_1)\land X_0(x_2)\land  X_1(x_3) \;\land\\
&(\forall y)[X_T(y)\limp\\
&  \hphantom{(\forall y)[}(
x\tmiabove y \liff (y\eq x_1\lor y\eq x_2\lor y\eq x_3)\;\land\\
&  \hphantom{(\forall y)[(}x_1\tmparent y\liff (y\eq x_2\lor y\eq x_3)\;\land\\
&  \hphantom{(\forall y)[(}\neg x_2\tmparent y\land\neg x_3\tmparent y\;\land\\
&  \hphantom{(\forall y)[(}x_2\tmileft y\liff y\eq x_3\;\land\\
&  \hphantom{(\forall y)[(}\neg x_3\tmileft y\quad)\qquad ]
\end{array}
\]
We can then require every node in $X_T$ to be licensed by some tuple in
$\sA$ by requiring it to satisfy $\bigvee_i[\varphi_i(x)]$, the disjunction of
such formulae for all tuples in $\sA$.
All that remains is to require the root to be labeled with an accepting state
and to ``hide'' the states by existentially binding them:
\[
\begin{array}{ll}
(\exists X_q)_{q\in Q}(\forall x)[\hspace*{-0.5em}&
(X_T(x)\limp\bigvee_i[\varphi_i(x)])\;\land\qquad\\
\multicolumn{2}{r}
{(\neg(\exists y)[y\tmiabove x]\limp\bigvee_{q\in Q_0}[X_q(x)])\,].}
\end{array}
\]
It is not hard to show that a $\Sigma$-labeled $3$-TM $\sT$ corresponds
to a satisfying assignment for this formula iff there is a run of $\sA$
on $\sT$ which assigns an accepting state to the root.

The proof that every set of trees definable in wSnT3 is recognizable, while a
little more involved, is essentially a lift of the proofs of Doner
\shortcite{doner70} and Thatcher and Wright~\shortcite{ThaWri68}.  The initial
step is to 
show that every formula in the language of wSnT3 can be reduced to equivalent
formulae in which only set variables occur and which employ only the predicates
$X\subseteq Y$ (with the obvious interpretation) and $X\tmpred{i} Y$
(satisfied iff $X$ and $Y$ are both singleton and the sole element of $X$
stands in the appropriate relation to the sole element of $Y$).  We can define,
for instance,
\[
\begin{array}{lr}
\multicolumn{2}{c}{\f{Empty}(X) \equiv
(\forall Y)[Y\subseteq X\limp X\subseteq Y]}\\[1ex]
\mbox{and}\\[1ex]
\multicolumn{2}{l}{\quad\f{Singleton}(X) \equiv}\\ 
&(\forall Y)[Y\subseteq X\limp(\f{Empty}(Y)\lor X\subseteq Y)]
\end{array}
\]
Then $x\tmpred{i}y$ becomes
\[
\f{Singleton}(X)\land\f{Singleton}(Y)\land X\tmpred{i}Y.
\]
It is easy to
construct $3$-TM automata (over the alphabet $\power(\set{X,Y})$) which accept
trees encoding satisfying assignments for these atomic formulae.  
For example, assignments satisfying $X\tmiabove Y$ in $\sfT^3_2$ are in
$\sA(2)$ for $\sA$: 
\[
\begin{array}{ll}
\tup{\emptyset,0,\sT}, & \sT\in\set{\eps,\littletree{0}{0}{0}},\\
\tup{\set{Y},1,\sT}, & \sT\in\set{\eps,\littletree{0}{0}{0}},\\
\tup{\set{X},2,\sT}, & 
 \sT\in\set{\littletree{1}{0}{0},\littletree{0}{1}{0},\littletree{0}{0}{1}},\\
\tup{\emptyset,2,\sT}, &
  \sT\in\set{\littletree{2}{0}{0},\littletree{0}{2}{0},\littletree{0}{0}{2}},\\
\tup{\sigma,3,\sT}, &\tow.
\end{array}
\]
The extension to arbitrary formulae (over these atomic formulae) can then be
carried out by induction on the structure of the formulae using the closure
properties of the recognizable sets.

\section{Defining TALs in wSnT3}\label{sec.apps}
The signature of wSnT3 is inconvenient for expressing linguistic constraints.
In particular, one of the strengths of the model-theoretic approach is the
ability to define long-distance relationships without having to explicitly
encode them in the labels of the intervening nodes.  We can extend the
immediate predecessor relations to relations corresponding to (proper)
\emph{above} (within the $3$-TM), \emph{domination} (within a tree), and
\emph{precedence} (within a set of siblings) using:
\[
\begin{array}{lr}
\multicolumn{2}{l}{x\tmbp{i}y\defiff x\not= y\land 
(\exists X)[X(x)\land X(y)\land}\\
&    (\forall z)[X(z)\limp(z\eq y\lor
     (\exists!z')[X(z')\land z\tmpred{i}z'])]].
\end{array}
\]
Which simply asserts that there is a sequence of (at least two) points linearly
ordered by $\tmpred{i}$ in which $x$ precedes $y$.\footnote{This is partly a
consequence of the fact that assignments to $X$ are required to be finite.}

To extend these through the entire structure we have to address the fact that
the two dimensional yield of a $3$-TM is not well defined---there is nothing
that determines which leaf of the tree expanding a node dominates the subtree
rooted at that node.  To resolve this, we extend our structures to include a
set $H$ picking out exactly one head in each set of siblings, with the ``foot''
of a tree being that leaf reached from the root by a path of all heads.
Given $H$, it is possible to define $\tmpdom$ and $\tmleft$, variations of
dominance and precedence\footnote{Of course $\tmpabove$ is just $\tmbabove$.} 
that are inherited by substructures in the appropriate way.  Let:
\[
\begin{array}{rcl}
\multicolumn{3}{l}{\quad
\f{Spine}_2(x) \defiff \f{H}(x)\land}\\
&& (\forall y)[y\tmbdom x \limp 
  (H(y)\lor \neg(\exists z)[z\tmparent y])\\[0.5ex]
\multicolumn{3}{l}{\mbox{and}}\\[0.5ex]
\multicolumn{3}{l}{\quad
  x\tmrtp{i} y \defiff x\tmptp{i} y\lor x\eq y.}\\[0.5ex]
\multicolumn{3}{l}{\mbox{Then}}\\
\multicolumn{3}{l}{\quad
  x\tmpdom y \defiff }\\
&& (\exists x',y')[
  x'\tmabove x\land y'\tmabove y\land x'\tmbdom y' \land\quad\qquad\\
\multicolumn{3}{r}{
  (\forall z)[ (x'\tmpabove z\land z\tmabove x)
                 \limp \f{Spine}_2(z)]\; ]
}\\
\multicolumn{3}{l}{\mbox{and}}\\
\multicolumn{3}{l}{\quad
x\tmleft y \defiff}\\
&& (\exists x',y')[x'\tmdom x\land y'\tmdom y\land
   x'\tmbleft y']
\end{array}
\]
At the same
time, it is convenient to include the labels explicitly in the structures.
A headed $\Sigma$-labeled $3$-TM, then, is a structure:
\[
\tup{T,\tmpred{i},\tmbp{i},\tmptp{i},H,P_\sigma}_{1\leq i\leq3,\,
\sigma\in\Sigma},
\]
where $T$ is a rooted, connected subset of $T^3_n$ for some $n$.

With this signature it is easy to define the set of $3$-TM that captures a TAG
in the sense that their $2$-dimensional yields---the set of maximal points wrt
$\tmpabove$, ordered by $\tmpdom$ and $\tmleft$---form the set of trees derived
by the TAG. Note that obligatory (OA) and null (NA) adjoining constraints
translate to a requirement that a node be \mbox{(non-)}maximal wrt $\tmpabove$.
In 
our automata-theoretic interpretation of TAGs selective adjoining (SA)
constraints are encoded in the states.  Here we can express them directly: 
a constraint specifying the modifier trees which may adjoin to an N node, for
instance, can be stated as a condition on the label of the root node of trees
immediately below N nodes.  

In general, of course, SA constraints depend not only on the attributes (the
label) of a node, but also on the elementary tree in which it occurs and its
position in that tree.  Both of these conditions are actually expressions of
the local context of the node. Here, again, we can express such conditions
directly---in terms of the relevant elements of the node's neighborhood.
At least in some cases 
this seems likely to allow for a
more general expression of the constraints, abstracting away from the
irrelevant details of the context.

Finally, there are circumstances in which the primitive locality of SA
constraints in TAGs is inconvenient.  
Schabes and Shieber~\shortcite{SchShi94}, for
instance, suggest allowing multiple adjunctions of modifier trees to the same
node on the grounds that selectional constraints hold between the modified node
and each of its modifiers but, if only a single adjunction may occur at the
modified node, only the first tree that is adjoined will actually be local to
that node.  They point out that, while it is possible to pass these constraints
through the tree by encoding them in the labels of the intervening nodes, such
a solution can have wide ranging effects on the overall grammar.  As we noted
above, the expression of such non-local constraints is one of the strengths of
the model-theoretic approach.  We can state them in a purely natural way---as a
simple restriction on the types of the modifier trees which can occur below (in
the $\tmpabove$ sense) the modified node.

\section{Conclusion}
We have obtained a descriptive characterization of the TALs via a 
generalization of existing characterizations of the CFLs and regular
languages.  These results extend the scope of the model-theoretic tools for
obtaining language-theoretic complexity results for constraint- and 
principle-based theories of syntax to the TALs and, carrying the generalization
to arbitrary dimensions, should extend it to cover a wide range of mildly
context-sensitive language classes.  Moreover, the generalization is natural
enough that the results it provides should easily integrate with existing
results employing the model-theoretic framework to illuminate
relationships between theories.  Finally, we believe that this characterization
provides an approach to defining TALs in a highly flexible and theoretically
natural way. 

%\bibliography{tmacl}

\end{document}